\documentclass[10pt,twocolumn,letterpaper,pagebackref,breaklinks,colorlinks,allcolors=cvprblue]{article}

\usepackage[pagenumbers]{cvpr} 

\definecolor{cvprblue}{rgb}{0.21,0.49,0.74}
\usepackage{hyperref}
\hypersetup{colorlinks,allcolors=cvprblue}

\usepackage{multicol}

\usepackage{amsmath}
\usepackage{amssymb}
\usepackage{gensymb}
\usepackage{nicefrac}
\usepackage{fontawesome}
\usepackage{pifont}
\usepackage{textcomp}
\usepackage{verbatim}
\newcommand{\cmark}{\ding{51}}%
\newcommand{\xmark}{\ding{55}}%

\usepackage{booktabs}
\usepackage{colortbl}
\usepackage{array}
\usepackage{ragged2e}
\usepackage{tabularx}
\usepackage{makecell}
\usepackage{multirow}
\usepackage{xspace}

\newcommand{\greencheck}{{\color{Green}\cmark}\xspace}
\newcommand{\green}{\cellcolor{Green!12.5}\greencheck}
\newcommand{\yellowcheck}{{\color{YellowOrange}(\cmark)}\xspace}
\newcommand{\yellow}{\cellcolor{YellowOrange!12.5}\yellowcheck}
\newcommand{\redcheck}{{\color{red}\xmark}\xspace}
\newcommand{\red}{\cellcolor{red!12.5}\redcheck}

\usepackage{enumitem}
\setlist{nosep}
\renewcommand{\paragraph}[1]{\par\vspace{2pt plus 1pt minus 1pt}\noindent{\bfseries #1\enspace}}
\def\paperID{4777} 
\def\confName{CVPR}
\def\confYear{2025}

\title{Tolerance-Aware Deep Optics}

\author{
    Jun Dai\textsuperscript{1} \hfill \quad
    Liqun Chen\textsuperscript{1}\footnote{test} \hfill \quad
    Xinge Yang\textsuperscript{2} \hfill \quad
    Yuyao Hu\textsuperscript{1} \hfill \quad
    Jinwei Gu\textsuperscript{3} \hfill \quad
    Tianfan Xue\textsuperscript{4,1}
    \\[0.4em]
    \textsuperscript{1}Shanghai AI Laboratory \quad
    \textsuperscript{2}KAUST \\
    \textsuperscript{3}NVIDIA \quad
    \textsuperscript{4}The Chinese University of Hong Kong
}
\begin{document}
\twocolumn[{%
\renewcommand\twocolumn[1][]{#1}%
\maketitle
\begin{center}
    \centering
    \captionsetup{type=figure}
    \includegraphics[width=1.0\textwidth,height=7.8cm]{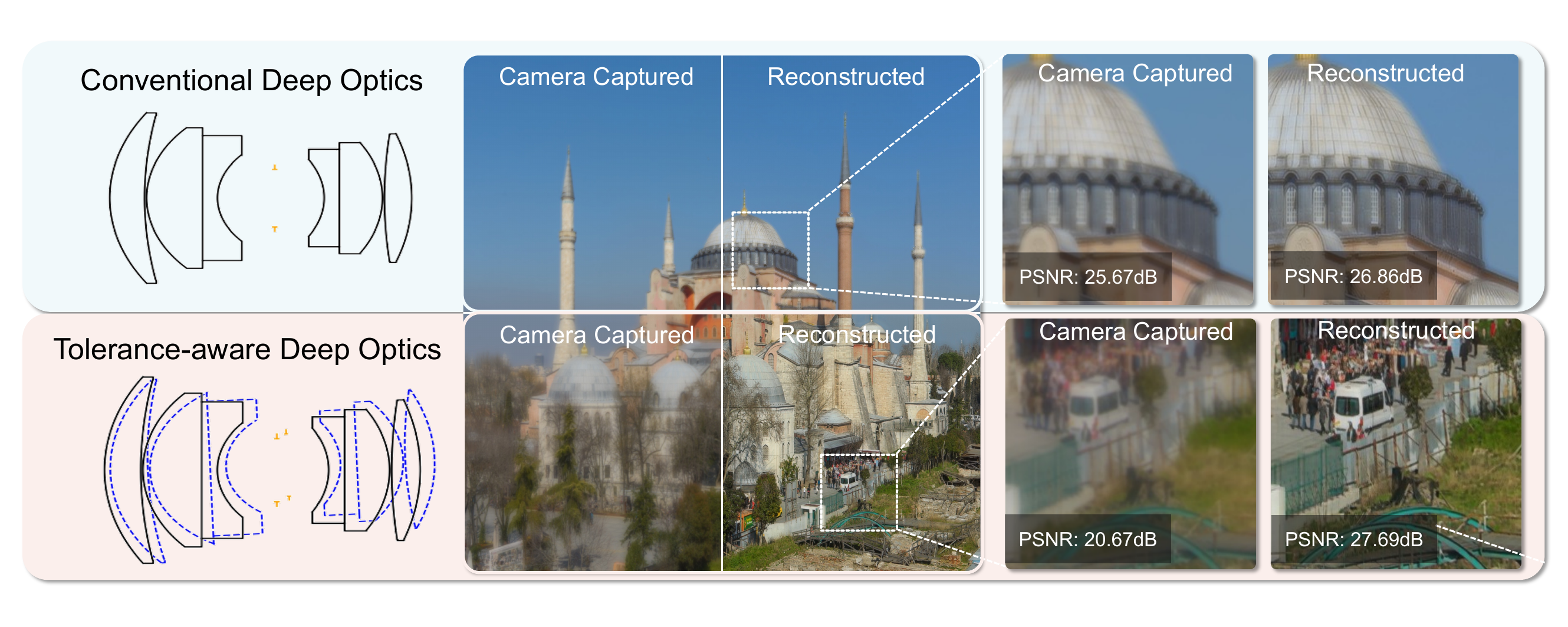}
    \captionof{figure}{When subjected to random tolerances perturbations, the performance of conventional deep optics degraded severely, whereas deep optics with our tolerance-aware optimization maintained excellent computational imaging performance. \emph{Camera Captured} and \emph{Reconstructed}, representing the simulated camera imaging result and the reconstructed result after decoder, respectively.}
\end{center}%
}]
\begin{abstract}
\footnotetext{Corresponding author. E-mail: chenliqun@pjlab.org.cn}Deep optics has emerged as a promising approach by co-designing optical elements with deep learning algorithms. However, current research typically overlooks the analysis and optimization of manufacturing and assembly tolerances. This oversight creates a significant performance gap between designed and fabricated optical systems. To address this challenge, we present the first end-to-end tolerance-aware optimization framework that incorporates multiple tolerance types into the deep optics design pipeline. Our method combines physics-informed modelling with data-driven training to enhance optical design by accounting for and compensating for structural deviations in manufacturing and assembly. We validate our approach through computational imaging applications, demonstrating results in both simulations and real-world experiments. We further examine how our proposed solution improves the robustness of optical systems and vision algorithms against tolerances through qualitative and quantitative analyses. Code and additional visual results are available at \href{https://openimaginglab.github.io/LensTolerance/}{openimaginglab.github.io/LensTolerance}.
\end{abstract}    
\section{Introduction}
\label{sec:intro}
The concept of deep optics~\cite{yang2024curriculum, Sitzmann_2018, sun2021end} has garnered significant attention for the joint optimization of optical systems and downstream vision algorithms. In a deep optics framework, the differentiable optical simulation model~\cite{sitzmann2018end, yang2024curriculum, sun2021end} and the downstream image reconstruction algorithm~\cite{sun2020learning, klotz2025minimalist} are optimized jointly in an end-to-end manner~\cite{li2021end, sun2021end}. This approach enables the design of novel optical systems that better support downstream tasks. End-to-end optical design~\cite{sitzmann2018end, peng2019learned, cai2024phocolens, yang2024curriculum,sun2021end} facilitates a tighter integration between hardware and software to achieve design objectives, with promising example applications observed across various vision tasks, including hyperspectral imaging~\cite{li2022quantization}, extended depth-of-field imaging~\cite{yang2024curriculum}, high-dynamic-range (HDR) imaging~\cite{sun2020learning, metzler2020deep}, and encoded depth estimation~\cite{yang2023aberration, ikoma2021depth}. Among existing optical systems, refractive lenses are still most widely used in practical camera platforms~\cite{wang2022differentiable, cote2023differentiable, peng2019learned, li2021end, zhang2023large}.

However, most deep optics solutions~\cite{yang2023aberration, chang2019deep, sun2020learning} assume perfect manufacturing and assembly processing, without any \textbf{tolerance-aware optimization}. In mass production, manufacturing and assembly errors are unavoidable and they may cause performance degradation (\cref{sec:methods}) on both the optical and algorithm sides. On the optical side, this degradation leads to a \textbf{design-to-manufacturing gap}, see \cref{fig:PerbVisualization}, confining the quality of deep optics and limiting its applicability in the real world. On the algorithm side, they also hurt the performance network-based image processing, as error-free optics are assumed at network training time. Although there are several exploration from traditional tolerance optimization~\cite{oinen1990new, forse1996statistical, hu2015design}, to mitigate the performance gap between lens design and manufacturing, it solely considers the optical component. In an end-to-end optical design pipeline, this will disrupt the encoding and decoding relationship between the optics and algorithms (\cref{sec:results}), compromising the integrity of the original design. 

Therefore, in this work, we proposed the first deep optics learning framework that explicitly models the tolerances in manufacture and assembly. This approach allows us to sample tolerances using Monte Carlo sampling and to optimize these tolerances in an end-to-end manner, ensuring that the final design meets machining requirements while preserving the overall design performances. With this framework in place, we further define a more complete deep optics optimization flow that includes a tolerance optimization process, i.e., firstly a pre-training stage without considering tolerances, in order to obtain an initial design that meets the design objectives, and secondly, tolerance-aware optimization is used to optimize the initial design, which makes deep optics much more robust to tolerances while meeting design objectives. This more complete deep optics optimization flow greatly reduces the design-to-manufacturing gap.

Nonetheless, due to the random sampling of multiple tolerance patterns during tolerance-aware optimization, this makes it very unstable and difficult to rely exclusively on the loss functions of design objectives (\cref{sec:ablations}). Therefore, to mitigate the instability and difficulty of the tolerances optimization process, we novelly design the Point Spread Function (PSF) similarity loss to enhance the robustness of the optical system to tolerances by constraining the PSF variation. In addition, we utilize Spot loss to ensure that the imaging properties of the optical system remain within reasonable range by limiting the spot size of the optical system. By combining the two proposed losses, it is possible to complete the tolerance optimization stage stably and effectively without unduly degrading the performance of the pre-trained design.

To validate the accuracy of our tolerance modeling, we compare the ray tracing with random tolerances with Zemax~\cite{ZEMAX}, and the results show that our modeling accuracy is close to that of Zemax. Additionally, we compare the performance gap between deep optical designs with and without tolerance optimization under random tolerances, using computational imaging as a case study. Our results demonstrate that tolerance optimization can achieve over $2\text{dB}$ improvement in deblurring performance (\cref{tab:result}). At the same time, we apply tiny perturbations to simulate tolerance perturbations during the actual images acquisition, and illustrate the effectiveness of our proposed tolerance-aware optimization by comparing the quality of the reconstructed images in the real-world experiment.

\section{Related Work}
Our work addresses an important problem - tolerances optimization - that has been neglected in previous deep optics. Here we review recent progress on tolerances optimization and deep optics, and highlight the main differences between existing approaches and our proposed method.

\subsection{Tolerances Analysis and Optimization}
Reduces the sensitivity of the optical design to tolerances and brings it in line with machining accuracy is a long standing problem in optical design. Traditional optical design typically involves a two-stage process \cite{laikin2018lens}, where the optical system is initially designed to fulfill the design objectives, followed by fine-tuning to align with manufacturing accuracy. The root-mean-square (RMS) or peak-to-valley (PV) values of the optical design are used to evaluate the sensitivity of optical component tolerances, as provided by conventional tolerance analysis theory~\cite{ni2019description, maksimovic2016optical, hu2015design}. The tolerance theory is a probabilistic statistical framework based on Monte Carlo methods~\cite{oinen1990new, forse1996statistical}, making it suitable for the tolerance analysis of mass-produced optical components. Zemax software~\cite{ZEMAX}  randomly samples tolerances, analyzes the sensitivity of design parameters to these tolerances through changes in design performance, and adjusts the parameters accordingly~\cite{Synopsys}.

In the context of deep optics for refractive lenses based on geometric ray tracing models, by measuring the actual Spatial Frequency response (SFR), Chen \textit{et al.} further re-calibrated the fabricated optical design parameters that were affected by the tolerances and  improved the accuracy of the imaging simulation in deep optics~\cite{chen2022computational, chen2021optical}. Zhou \textit{et al.} built on their work by adjusting the spacing between some lenses to align the simulated point spread function (PSF) with the actual measured PSF. At the same time, they employed a more powerful Mamba model~\cite{gu2023mamba} as the decoder, thereby reducing the sensitivity of deep optics to tolerances~\cite{zhou2024optical}. However, the aforementioned works focus on calibrating an already machined optical system rather than end-to-end optimization of design parameters, and they do not address sensitivity to tolerances at the design stage. Closest to our work are Zheng \textit{et al.}~\cite{zheng2023neural} using fabrication simulator to simulate degradation due to fabrications in computational lithography and Li \textit{et al.}~\cite{li2022quantization} in deep optics considering Diffractive Optical Elements (DOE) quantization errors due to fabrications, however, they both are not focus on lens, we list all comparisons in \cref{tab:related_work}.

\subsection{Deep Optics}
Deep optics is an emerging field that jointly design and optimize the optical systems and vision algorithm using deep learning~\cite{wetzstein2020inference}. Sitzmann \textit{et al.}~\cite{sitzmann2018end} utilized this approach to design an optical element that enhances image quality. Subsequently, more works have used this method to design optics for image enhancement~\cite{dun2020learned, metzler2020deep, sun2020learning, chakravarthula2023thin} and depth estimation~\cite{chang2019deep, haim2018depth, wu2019phasecam3d, baek2021single}. And similar approach has been taken to design imaging lenses using differentiable ray tracing~\cite{sun2021end, wang2022differentiable, yang2024curriculum}, and differentiable proxy functions~\cite{tseng2021differentiable, yang2023aberration}.~\citet{tseng2021neural} used this technique to design a metasurface lens with improved image quality. In each of these works, a differentiable model for camera's optics is incorporated into a neural network, and the optics is designed by training the network for the specific task.

However, current deep optics models do not consider the impact of tolerances in the simulation model. This problem causes the mismatch between the designed and the fabricated lens system, which degrades reconstruction quality in physical systems. In contrast, our tolerance-aware optimization fix this mismatch by directly modeling common tolerances.

\begin{table}[t]  
    \setlength{\tabcolsep}{0em}  
    \renewcommand{\arraystretch}{1.2}
    \centering  
    \footnotesize  
    \caption{  
        Comparison of related work on the tolerances optimization of deep optics and computational optics, where each criterion is fully~\greencheck, partially~\yellowcheck, or not~\redcheck met.  
        See text for explanations.  
        }  
\begin{tabularx}{\linewidth}{m{0.25\linewidth}XXXXX}
    \toprule
    &
    {\footnotesize Chen \textit{et al.}~\cite{chen2022computational}} &  
    {\footnotesize Zhou \textit{et al.}~\cite{zhou2024optical}} &  
    {\footnotesize Zheng \textit{et al.}~\cite{zheng2023neural}} &  
    {\footnotesize Li \textit{et al.}~\cite{li2022quantization}} &  
    {\footnotesize Ours} \\
    \midrule
    \textbf{Optics Type} & Lens & Lens & DOE & DOE & Lens \\ 
    \midrule
     \scalebox{0.85}{Optimize Parameters} & 
     \red & 
     \yellow & 
     \green & 
     \green &
    \green \\
    Optics \& Decoder & 
    \red & 
    \green & 
    \green & 
    \green &
    \green \\
    Explicit modeling & 
    \green & 
    \green & 
    \red & 
    \green &
    \green \\
    \scalebox{0.9}{Improve Robustness} & 
    \red & 
    \red & 
    \green & 
    \green &
    \green \\
    \bottomrule
\end{tabularx}
  
    \label{tab:related_work}  
\end{table}

\section{Methods}
\label{sec:methods}
In this section, we introduce our tolerance-aware deep optic optimization. First, in \cref{sec:3.1}, we introduce the definition of tolerance optimization in term of deep optics. Then, in \cref{sec:3.2}, we explicitly model four common tolerances, decentering, tilt, curvature and central thickness errors in the differentiable imaging process. Finally, in \cref{sec:3.3} and \cref{sec:3.4}, we introduce two novel losses and present a more completed design flow for deep optics, which includes the pre-training process without tolerances and the tolerance-aware optimization that accounts for actual errors.

\subsection{Tolerance Optimization for Deep Optics}
\label{sec:3.1}
In traditional optical design, tolerance optimization involves optimizing the design parameters to ensure that the degradation resulting from manufacturing or assembly errors remains within an acceptable range. Specifically, we assume the actual lens parameters $\theta_i$ at manufacturing time follows independent Gaussian distributions centered around designed parameters, $\mathcal{N}(\tilde{\theta_i}, \tilde{\sigma_{i}^2})$, where $\tilde{\theta_i}$ is the design parameters without any tolerances and $\tilde{\sigma_{i}^{2}}$ is the variance of tolerance range. Then, within given ranges of tolerances $t$, we need to find an optimal lens parameters $\boldsymbol{\theta_{\text{opts}}}$, like curvature or spacing distance, where the imaging quality degradation is bounded as:
\begin{equation}
\boldsymbol{\theta_\text{{opt}}^{*}} = \mathop{\arg}\mathop{\max}_{\boldsymbol{\theta_{\text{opt}}}} \mathcal{P}(\mathrm{Y} \negthinspace > \negthinspace \mathrm{Y_0} \negthinspace \mid \negthinspace {\boldsymbol{\theta_{opt}}}, \negthinspace t),
\label{eq:1}
\end{equation}
where  $Y$ is the actual imaging quality of a lens, $Y_0$ is the preset performance threshold, and $\mathcal{P}(\mathrm{Y} \negthinspace > \negthinspace \mathrm{Y_0})$ is the probability that the lens performance is within the designed range.

Moving to the deep optical design, the main difference is that we also need to consider the downstream image processing networks. Therefore, \cref{eq:1} is modified to:
\begin{equation}
\boldsymbol{\theta_{\text{opt}}^{*}},\boldsymbol{\theta_{\text{net}}^{*}}= \mathop{\arg}\mathop{\max}_{\boldsymbol{\theta_{\text{opt}}},\boldsymbol{\theta_{\text{net}}}} \mathcal{P}(\mathrm{Y}\negthinspace >\negthinspace \mathrm{Y_0}\negthinspace \mid \negthinspace{\boldsymbol{\theta_{\text{opt}}}},\negthinspace{\boldsymbol{\theta_{\text{net}}}}, \negthinspace t),
\label{eq:2}
\end{equation}
where $\boldsymbol{\theta_{\scalebox{0.5}{net}}}$ is the parameters from downstream algorithms, like the parameters of neural networks. Also, $Y$ is the performance of the downstream task, instead of imaging quality.

\begin{figure}[t]
  \centering
   \includegraphics[width=0.95\linewidth, height=3.7cm]{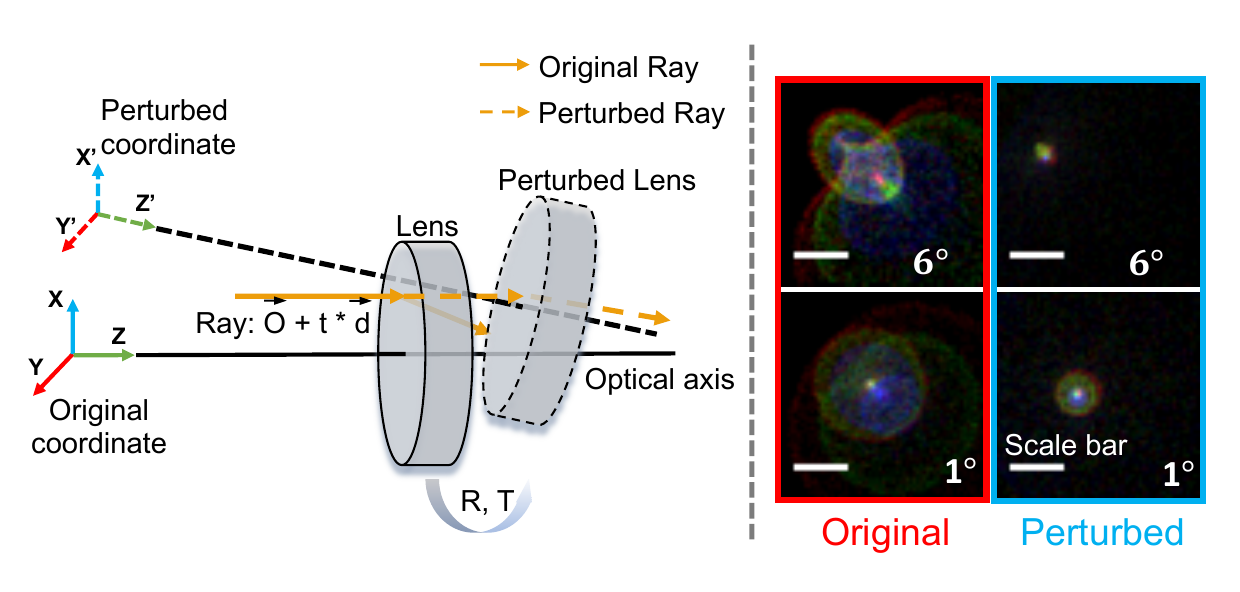}

   \caption{The tilt, decentration and central thickness tolerances can be modeled as {\bf{R}}otation and {\bf{T}}ranslation of Lens. The spatial transformations of Lens can be equivalent to the opposite spatial transformations of ray's coordinate. The PSFs of the lens system change drastically due to tolerances. Scale bar: 15$\mu m$.}
   \label{fig:PerbVisualization}
\end{figure}

\subsection{Modeling Tolerances in Ray Tracing}
\label{sec:3.2}
Due to the independent tolerances of each lens and the interdependence among various types of tolerances, employing a global a priori approach to make the optical system robust against tolerances is challenging~\cite{mcguire2006designing}. To address this, we propose explicitly modeling tolerances in every iteration of ray tracing where most of previous imaging simulation pipeline do not, allowing the optics and computational decoder to be aware of the effects of tolerances. Specifically, we use a forward ray tracing algorithm \cite{kolb1995realistic, wang2022differentiable}, taking into accounts of four common lens tolerances: decentration, tilt, central thickness, and curvature errors. We perturb the original lens by these tolerances in a differentiable way and integrate them into the previous joint optimization framework.

The tolerances arising from decentration, tilt, and central thickness errors can be represented as translations and rotations of the lens surfaces, as illustrated in \cref{fig:PerbVisualization}. Decentration and central thickness tolerances correspond to rigid translations of the lens in the plane perpendicular to the optical axis and along the optical axis, respectively. In contrast, the tilt tolerance can be represented as a rigid rotation of the lens, with the transformation detailed in \cref{eq:transformation}. However, to simplify the verification of boundary conditions during ray tracing, we convert the rigid lens transformation into an equivalent coordinate transformation. For curvature error tolerance, a curvature offset $\Delta c$ is applied in the forward ray tracing process.

In summary, to solve for the intersections of the rays with the lens during the ray tracing process, we transform the coordinates according to randomly perturbation. Then, after solving the intersection of the rays on the lens and the refraction process, we re-convert the coordinates to the original global coordinate system.
\begin{align}
\begin{matrix} 
[\mathbf{P^{\prime}}, \mathbf{d^{\prime}}]
\end{matrix} = \mathbf{R} \cdot
\begin{matrix} [\mathbf{P}+\mathbf{\Delta T}, \mathbf{d} ]
\end{matrix},
\label{eq:transformation}
\end{align}
where $\mathcal{\boldsymbol{P}}$ and $\mathcal{\boldsymbol{P^{\prime}}}$ are the original and perturbed ray origins, and $\mathcal{\boldsymbol{d}}$ and $\mathcal{\boldsymbol{d^{\prime}}}$ are the original and perturbed ray directions; $\mathcal{\boldsymbol{\Delta T}}$ is translation vector and $\mathcal{\boldsymbol{R}}$ is the rotation matrix.

\begin{figure*}[tb] \centering
    \includegraphics[width=\textwidth,height=0.43\textwidth]{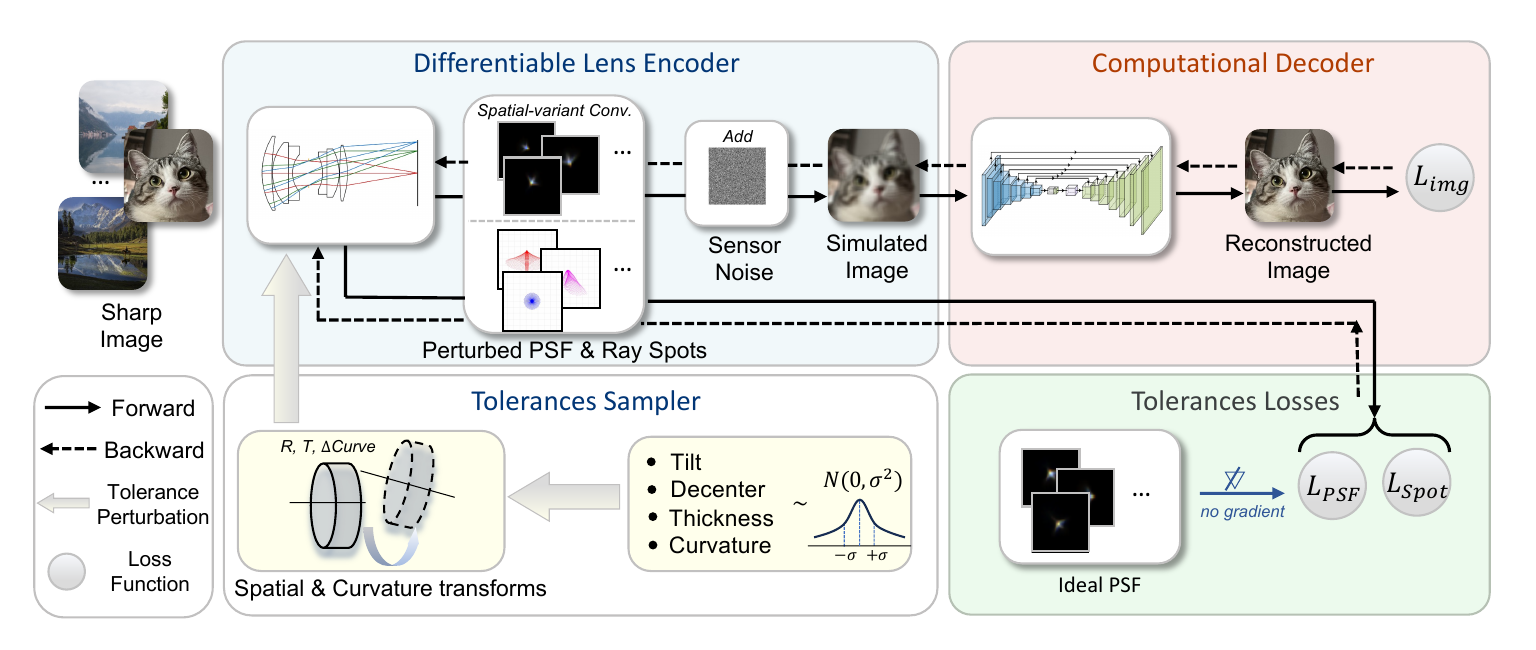}
    \caption{{\bfseries Tolerance-aware optimization for deep optics}. We integrate tolerances into differentiable ray tracing. Every kind of tolerances are randomly sampled from its distribution $\mathcal{N}($0$, \tilde{\sigma_{i}^2})$, use ray tracing with tolerances to render perturbed spatially-variant PSF maps and simulated the imaging results by spatially-variant convolution, then noise is added to simulate sensor-captured images. These images are then passed to a computational decoder for reconstruction. During forward simulation, we track gradients of optical parameters, We can subsequently back-propagate the errors from either the reconstruction images quality and tolerance loss terms. The framework jointly optimizes the optics and the computational decoder in a tolerance-aware manner.} 
    \label{fig:pipeline}
\end{figure*}

\subsection{Tolerance-aware Deep Optics Design}
\label{sec:3.3}
Explicitly modeling tolerances enables us to achieve tolerance-aware co-design. However, the random sampling of tolerances results in training instability, which is especially pronounced during the initial stages of deep optics optimization. Hence, we introduce a deep optics design flow that incorporates tolerance optimization and is easier to train: firstly, we employ the conventional deep optics training approach without accounting for tolerance effects, focusing solely on task performance as the design objective; secondly, once the design meets the design objectives, we conduct tolerance-aware optimization to keep its performance after fabrication.

For second tolerances optimization stage, we explicitly modeling tolerances into ray tracing process, then we use ray tracing with tolerances to implement tolerance-aware optimization, jointly optimize both the optics and computational decoder, see in \cref{fig:pipeline}, to improve the robustness of deep optics design. In each iteration, we randomly sample $N=64$ tolerances patterns and conduct forward ray tracing with sampled tolerance to render the PSF map for full field of views which contains the effects of all sampled tolerances, and use the mixed PSF map to render the sharp images by spatially-variant convolution \cite{cai2024phocolens, chen2021extreme}.

\subsection{Spot Loss and PSF Similarity Loss}
\label{sec:3.4}
Since deep optics contains, optics and decoder, the difference between the number of parameters and the significance of the parameters of the two is very large, and due to the tolerance-aware optimization when randomly sampling a variety of tolerance modes, simply using the general image quality loss, such as mean-square-error loss, VGG Loss \cite{johnson2016perceptual} and TV loss \cite{chambolle2010introduction}, the overall training difficulty is is very large, see \cref{tab:loss_comparison}. In this paper, we use a basic image quality loss, $\mathcal{L}_{img} = \lambda_{vgg}\cdot\mathcal{L}_{vgg} + \lambda_{tv}\cdot\mathcal{L}_{tv} + \lambda_{mse}\cdot\mathcal{L}_{mse}$, and $\lambda_{vgg}, \lambda_{tv}$ and $\lambda_{mse}$ are set in $0.001, 0.01$ and $0.1$.
\begin{align}
\mathcal{L}_{total} &= \lambda_{Spot}\cdot\mathcal{L}_{Spot} + \lambda_{PSF}\cdot\mathcal{L}_{PSF} + \mathcal{L}_{img}, \label{eq:Loss-total}
\end{align}
In order to improve stability the tolerances optimization for deep optics, we introduce two novel loss functions, $\mathcal{L}_{Spot}$ and $\mathcal{L}_{PSF}$, in addition to basic image quality losses, $\mathcal{L}_{img}$. We use the spot loss, $\mathcal{L}_{Spot}$ to constrain the spot size of lens which represent the overall imaging quality. 
And the PSF play the role of bridging the optics with the computational decoder, the tolerances are directly affect the PSF of lens, therefore we design $\mathcal{L}_{PSF}$ to constrain the change of PSF under random tolerances, the two kinds of losses expressed as \cref{eq:Loss-RMS} and \cref{eq:Loss-PSF1,eq:Loss-PSF2}. Noteworthy, the two new losses are directly decided by the lens, which are able to provide shortcuts for the backward gradients and improve the optimization for the lens parameters. In summary, we implement tolerance-aware optimization by loss functions, expressed \cref{eq:Loss-total}.
\begin{align}
\mathcal{L}_{Spot} &= \sum_{\scalebox{0.8}{$\lambda$}} \sum_{\scalebox{0.8}{$f$}} \frac{\|\mathbf{P}_{\lambda, f} - \overline{\mathbf{P}}_{\lambda, f}\|_2}{N_{rays}}, \label{eq:Loss-RMS} \\
\mathcal{L}_{PSF} &= 1.0 - \frac{\sum_{\scalebox{0.8}{$f $}} \sum_{\scalebox{0.8}{$\lambda$}} Sim_{\lambda, f}}{N_f N_{\lambda}}, \label{eq:Loss-PSF1}\\
Sim_{\lambda, \scalebox{0.8}{$f$}} &= \max\left( \mathrm{Conv}\left( \mathrm{PSF}_{Ideal}\mathrm{PSF}_{Perb} \right) \right), \label{eq:Loss-PSF2}
\end{align}
where $\mathbf{P_{\lambda, f}}$ and $\overline{\mathbf{P}}_{\lambda, f}$ are the position of every traced ray and averaged position of given wavelength and field of view (FoV). $N_{\text{rays}}, N_f$ and $N_{\lambda}$ are the number of sampled rays, FoVs and wavelengths. $\mathrm{Conv(\cdot)}$ represents $stride=1$ and $padding=K // 2$ convolution, and $K$ is the resolution size of a single PSF. $\lambda_{Spot}$ and $\lambda_{PSF}$ are depends on lens structure.

\section{Experiments}
\label{sec:results}
We conduct our experiments based on the open-source differentiable ray tracer DeepLens~\cite{yang2024curriculum,wang2022differentiable}.We firstly compare the ray tracing results of our framework with Zemax \cite{ZEMAX} to demonstrate the accuracy of imaging simulation with tolerances in \cref{sec:4.1}. And in \cref{sec:4.2} and \cref{sec:4.3}, we implement computational imaging task to validate the effectiveness of the tolerance-aware optimization in simulation and real-world levels. Finally, in \cref{sec:4.4}, we analyze how tolerance-aware optimization improve deep optics robustness to tolerances from both quantitative and qualitative perspectives.

\subsection{Tolerance-aware Differentiable Ray-tracing}
\label{sec:4.1}
Using the method described in \cref{sec:3.2}, we capture the deviations caused by random tolerance perturbations during the ray tracing process. This approach preserves the differentiability of the conventional deep optics pipeline, enabling accurate imaging simulations in the presence of tolerances. We demonstrate that the ray tracing with tolerances is quite accurate, compared the ray tracing results with Zemax \cite{ZEMAX}. We choose a classical optical design, Cooke Triplet, and randomly sample tolerances pattern to perturb the lens system. 

\begin{figure}[t]
  \centering
   \includegraphics[width=1.0\linewidth]{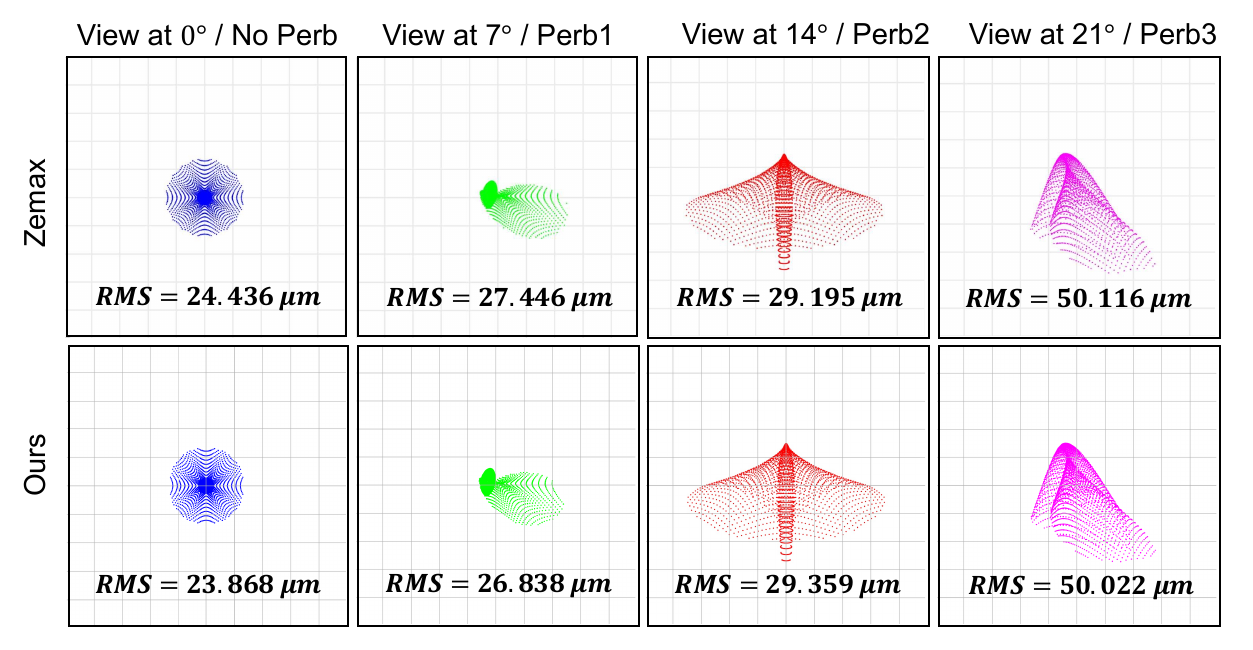}
   \caption{Spot diagrams and RMS spot sizes produced by our framework with and without perturbations are highly resemble those by Zemax, the tested lens is Cooke Triplet and for visualization, the scale bars are different.}
   \label{fig:ZEMAX}
\end{figure}

We visualize the ray tracing results come from Zemax \cite{ZEMAX} and ours, as shown in \cref{fig:ZEMAX}, and also calculate the root-mean-square spot sizes. The spot sizes and the overall rays distribution are nearly identical across each field-of-views, with errors $ < 1\mu m$, which shown that our ray tracing with tolerances is extremely accurate, thus it means that we are able to simulate the imaging process when lens encounter tolerances and remain differentiable manner of deep optics pipeline.

\begin{table*}[tbp]
    \centering
    \setlength{\tabcolsep}{.25em}
    \footnotesize
    \setlength{\aboverulesep}{1pt}
    \setlength{\belowrulesep}{1pt}
    \renewcommand*{\arraystretch}{0.75}
    \caption{Up: the numerical results of multiple (100 times) random sampled tolerances pattern test on DIV2K test datasets for without, with our tolerance-aware optimization and conduct tolerance optimization by Zemax \cite{ZEMAX}. Bottom: the manufacturing yield result based on 100 times results, \eg , $>90\%$ means that have $90\%$ confidence to get PSNR $>26.26dB$ for Lens1 after fabrications. }
    \renewcommand{\arraystretch}{1.3} 

\begin{tabularx}{\linewidth}{XccccccXccccc}
    \toprule
    \multicolumn{2}{c}{\multirow{2}{*}{\textbf{Optics}}} & & \multicolumn{3}{c}{\textbf{Test with tolerances}} & & \multicolumn{2}{c}{\multirow{2}{*}{\textbf{Optics}}} & & \multicolumn{3}{c}{\textbf{Test with tolerances}} \\
    \cmidrule{4-6} \cmidrule{11-13}
    \multicolumn{2}{c}{} & Spot$_\downarrow$($\mu$m) & PSNR$_\uparrow$ \cite{fardo2016formal} & SSIM$_\uparrow$ \cite{wang2004image} & LPIPS$_\downarrow$ \cite{zhang2018unreasonable} & & \multicolumn{2}{c}{} & Spot$_\downarrow$($\mu$m) & PSNR$_\uparrow$ & SSIM$_\uparrow$ & LPIPS$_\downarrow$ \\
    \midrule
    \multirow{3}{*}{\textbf{Lens1}}& w/ TOLR & 16.2 & \cellcolor{red!12.5} \textbf{29.61} & \cellcolor{red!12.5} \textbf{0.847} & 0.293 &  & \multirow{3}{*}{\textbf{Lens2}}& w/ TOLR & 38.7 & \cellcolor{red!12.5} \textbf{28.08} & \cellcolor{red!12.5} \textbf{0.850} & \cellcolor{red!12.5} \textbf{0.225} \\
    & w/o TOLR & 7.3 & 29.25 & 0.828 & \cellcolor{red!12.5}\textbf{0.268} & & & w/o TOLR & 14.2 & 25.75 & 0.735 & 0.334 \\
    & Zemax & \cellcolor{red!12.5} \textbf{6.2} & 29.14 & 0.824 & 0.270 &  &  & Zemax & \cellcolor{red!12.5} \textbf{9.7} & 23.58 & 0.779 & 0.248 \\
    \midrule
    \multirow{2}{*}{} & \multirow{2}{*}{} & \multicolumn{4}{c}{\textbf{Manufacturing Yield (PSNR)}}&  & \multirow{2}{*}{} & \multirow{2}{*}{} & \multicolumn{4}{c}{\textbf{Manufacturing Yield (PSNR)}} \\
    \cmidrule{3-6} \cmidrule{10-13}
     &  & $>90\%$ &  $>70\%$ &  $>50\%$ &  $>10\%$ &  & & &  $>90\%$ &  $>70\%$ &  $>50\%$ &  $>10\%$ \\
    \midrule
    \multirow{3}{*}{\textbf{Lens1}}& w/ TOLR & \cellcolor{red!12.5}\textbf{26.26} & \cellcolor{red!12.5}\textbf{29.18} & \cellcolor{red!12.5}\textbf{30.47} & 31.19 &  & \multirow{3}{*}{\textbf{Lens2}}& w/ TOLR & \cellcolor{red!12.5} \textbf{24.84} & \cellcolor{red!12.5}\textbf{27.03} & \cellcolor{red!12.5}\textbf{28.55} & 30.13 \\
    & w/o TOLR & 24.30 & 26.68 & 29.15 & 33.82& & & w/o TOLR & 21.94 & 24.26 & 26.40 & \cellcolor{red!12.5}\textbf{30.66} \\
    & Zemax & 24.13 & 26.55 & 29.08 & \cellcolor{red!12.5}\textbf{34.11} & &   & Zemax & 20.95 & 22.25 & 23.26 & 26.48 \\
    \bottomrule
\end{tabularx}
    \label{tab:result}
\end{table*}
\subsection{Tolerance-aware Evaluation by Simulation}
\label{sec:4.2}
To demonstrate that our tolerance-aware optimization provides better robustness and less degradation when subjected to tolerances, we compare the average deblurring results of the tolerance-aware deep optics design and the non-tolerance-aware counterpart under random tolerances, tested on the DIV2K dataset \cite{agustsson2017ntire}, shown in \cref{tab:result}. The design with tolerance-aware optimization, has better average deblurring performances when encounter tolerances, the average PSNR improved by more than $2\text{dB}$.

We also compare our tolerance-aware design with tolerance optimized by Zemax \cite{ZEMAX} which use Zemax to implement tolerances optimization merely by optical metrics alone, it shown that optimized by Zemax may incur the mismatch problem between optical and decoder part, thus lead the deblurring performance drop down severely (see Lens2 in \cref{tab:result}).

At last, we analyze the manufacturing yields of the three design approaches based on our test results, which showed that deep optics demonstrated higher manufacturing yields after tolerance-aware optimization, shown in \cref{tab:result}.

\begin{figure*}[tb] \centering
  \includegraphics[width=\textwidth,height=0.55\textwidth]{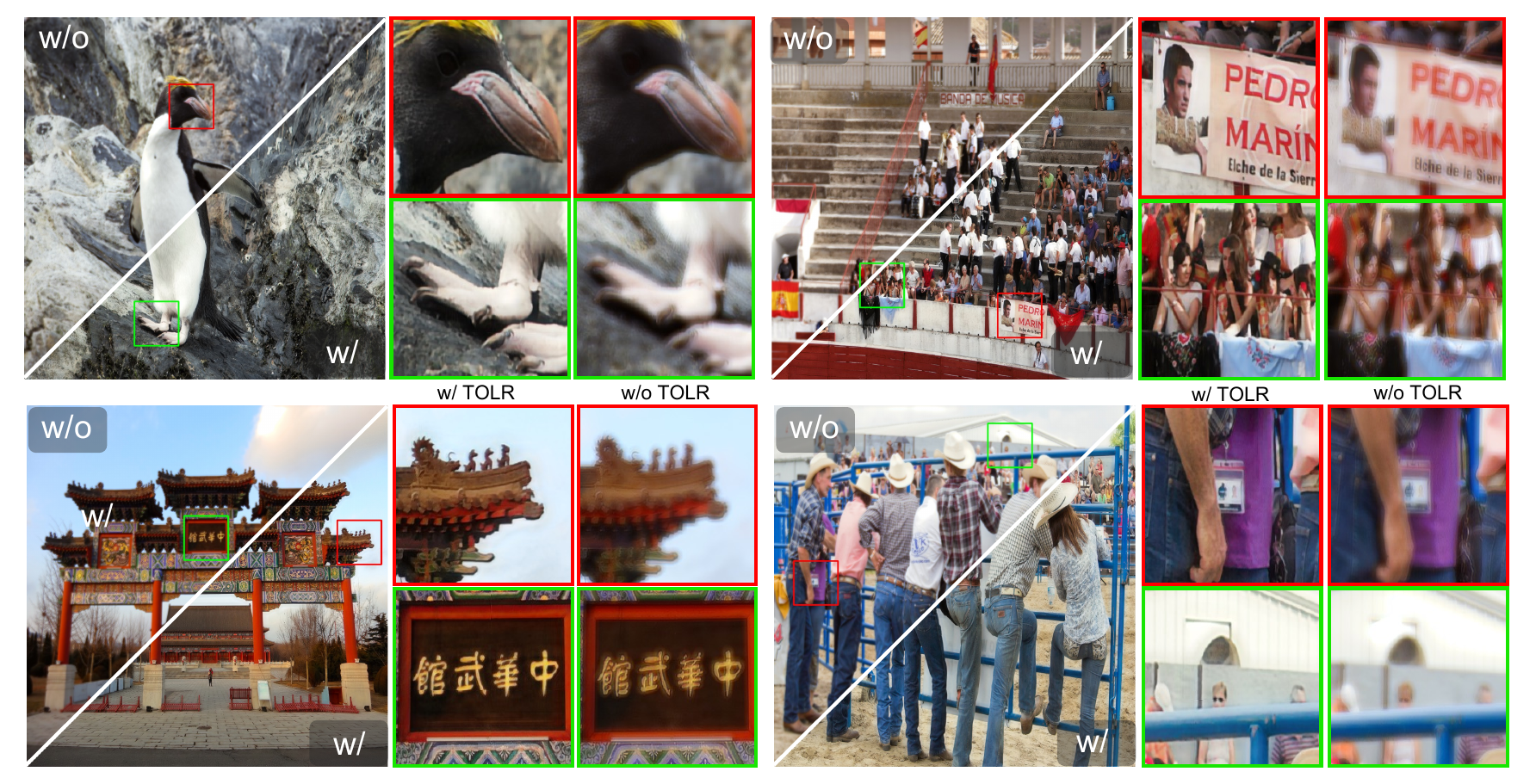}
  \caption{Deblurring results comparison of deep optics with and without tolerance-aware optimization under tolerances perturbations. The deep optics with tolerances optimization maintain better deblurring performances than its counterpart.} 
  \label{fig:Deblur}
\end{figure*}

\subsection{Real-world Experiment}
\label{sec:4.3}
\begin{figure*}[tb] \centering
  \includegraphics[width=\textwidth,height=0.42\textwidth]{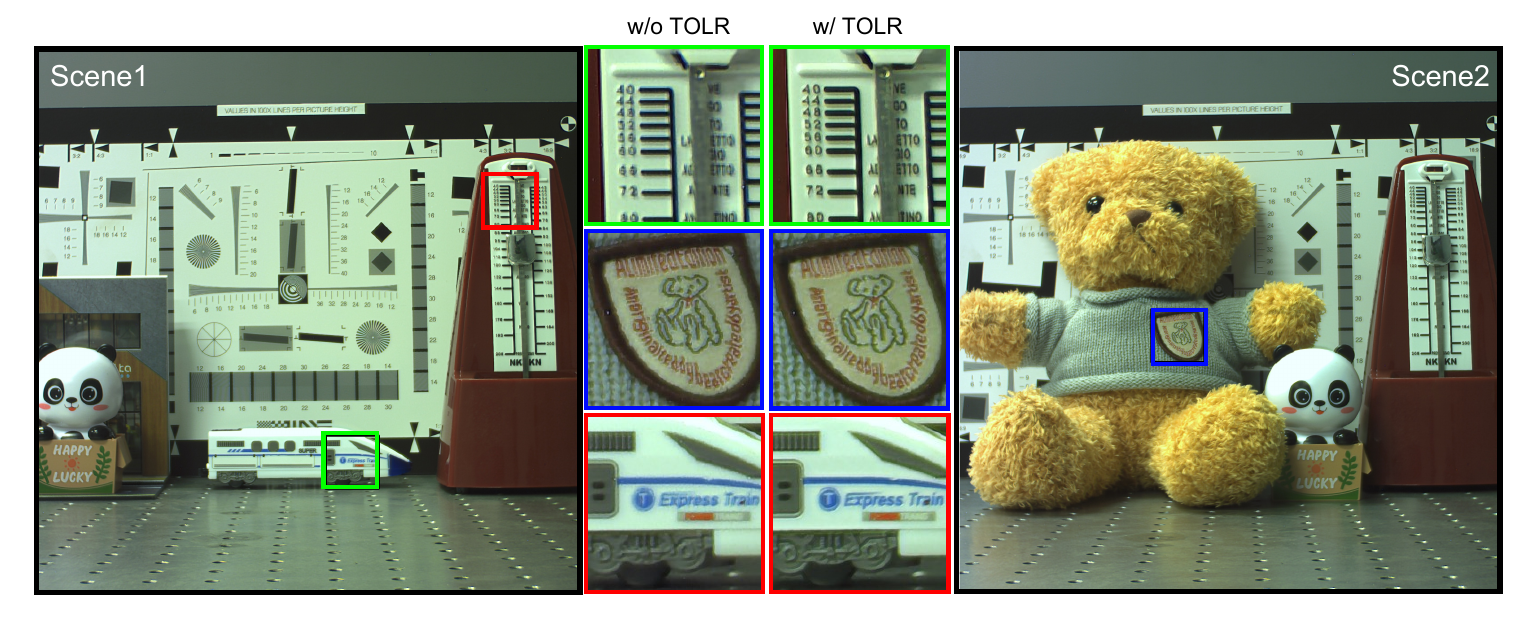}
  \caption{Physical experiment results demonstrating the robustness of our tolerance-aware deep optics. Despite random tolerance perturbations, our method maintains high deblurring performance in two distinct scenes. In contrast, the deep optics baseline without tolerance-aware optimization suffers from significant performance degradation.} 
  \label{fig:physical_exp_results}
\end{figure*}

To demonstrate the applicability of our method to real-world systems, we employ actual optical systems to acquire images and perform computational imaging. This shows that the tolerance-aware deep optics produces superior deblurring results in real-world which exist random tolerances. Specifically, we use an off-the-shelf industrial camera lens design to train and optimize the decoder both with and without tolerances optimization. We simulate tolerances by introducing slight perturbations during image acquisition and compare the quality of reconstructed images before and after optimizing the decoder for tolerances. This comparison highlights the enhanced resistance to potential tolerances provided by the optimized decoder, see \cref{fig:physical_exp_results}.

\subsection{Robustness Improvement of Deep Optics}
\label{sec:4.4}
The tolerance-aware optimization significantly enhances the robustness of the deep optics against potential tolerances. However, what specific factors contribute to this increased tolerances resistance following the optimization?

\begin{figure}[t]
  \centering
    \hspace{-5mm}
  \includegraphics[width=8.3cm, height=4.3cm]{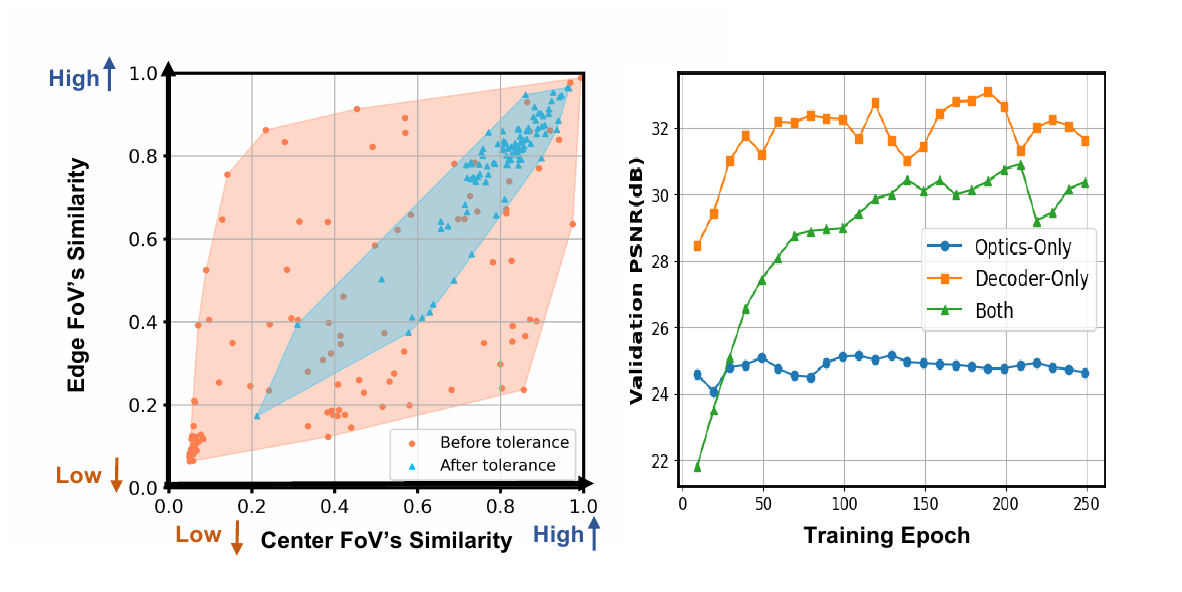}
   \caption{Qualitative analysis results for the optics (random test 50 times) and the validation PSNR values during training (test without tolerances).}
   \label{fig:qualitatively-analysis}
\end{figure}

To understand the underlying reasons, we qualitatively analyze changes in both the optics and decoder components after tolerance optimization. For the lens, we validate its tolerance robustness by comparing the degree of PSF variation in designs with and without tolerances optimization when affected by tolerances. By assessing how the PSFs change when subjected to random tolerances, using cosine similarity to quantify the similarity between the perturbed PSF and the ideal PSF, as shown in \cref{fig:qualitatively-analysis}. It is evident that the lens optimized with tolerance-aware can maintain the PSF within a relatively similar range, even when subject to tolerance disturbances. This greatly reduces the difficulty for the computational decoder to handle the changes of encoded patterns with tolerances.

\begin{figure}[t]
  \centering
  \includegraphics[width=8cm, height=3cm]{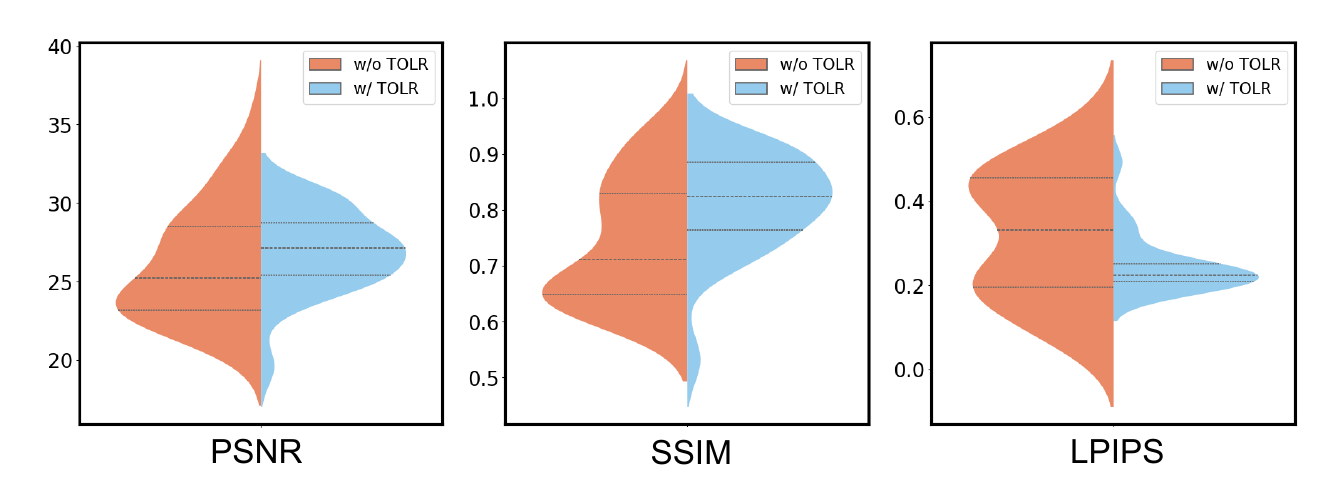}

   \caption{Qualitative analysis results for the decoder's robustness improvements. The fitted probability density distributions of 50 times of random tolerance tests of PSNR, SSIM and LPIPS. With and without tolerance optimization and for deep optics with tolerance-aware optimization is tested under larger range of tolerances.}
   \label{fig:qualitatively-analysis-decoder}
\end{figure}

Regarding the decoder, given the increased robustness of the lens post-optimization, we expand the range of tolerances to compare the decoder's ability to handle tolerance before and after tolerance optimization under equivalent conditions, as shown in \cref{fig:qualitatively-analysis-decoder}. The results demonstrate that with tolerance-aware optimization, the decoder become robust to potential tolerances. After tolerance-aware optimization, the lens can maintain PSF stability, and the decoder shows greater adaptability to tolerances. This significantly enhances the resistance of deep optics to manufacturing and assembly tolerances.
\section{Discussion and Ablation Study}
\label{sec:ablations}
We provide an in-depth analysis of our optimization setting, the impact of different loss functions, and the number of tolerances sampled during optimization. This analysis aims to offer more guidance for future work.

\subsection{Partially Tolerance Optimization}
In deep optics, the optimization of optical and decoder parameters may vary~\cite{tseng2021differentiable}. Tolerance-aware optimization adds further complexity by introducing random perturbations in each forward pass. To delve into the distinctions between these components, we selectively control the optimization: (a) optimizing the optics, and (b) only the computational decoder. Through experiments, we find that the deep optics achieves improvement of robustness only when the optics and decoder are jointly optimized simultaneously, shown in \cref{tab:partly-tolerance}. 
Optimizing only the decoder can enhance the decoder's tolerance-aware ability, but still fails to fully address the impact of random tolerances. Additionally, when optimizing the optics alone, the global influence of optical parameters on the subsequent decoder, along with the limited number of optical parameters, makes the optimization challenging and risks compromising the results of the initial pre-training stage. Only through the joint optimization of the optical system and decoder can we achieve a more stable optical system with enhanced pairing capability from the decoder, significantly improving the robustness of deep optics against potential tolerances.


\begin{table}
\centering
\caption{\emph{Optics/Decoder-only} means that only optimize the optics/decoder part of parameters during tolerance optimization and \emph{Both} means optimized both parts. The results are average of 100 times of random sampled tolerances experiments (excluding Spot Size).}
\begin{tabular}{c|c c c c}
\hline
 & Optics-only & Decoder-only & Both \\
\hline
PSNR$\uparrow$ & 22.06 & 25.94 & \bf{28.08} \\
SSIM$\uparrow$ & 0.376 & 0.734 & \bf{0.845} \\
LPIPS$\downarrow$ & 0.664 & 0.379 & \bf{0.225} \\
Spot Size ($\mu m$)$\downarrow$ & 34.2 & \textbf{14.2} & 38.7 \\
\hline
\end{tabular}
\label{tab:partly-tolerance}
\end{table}

\subsection{Analysis of Loss Function Impact}
To analyze the impact of the two loss functions, we conduct ablation studies on Spot loss and PSF loss. The experimental results are shown in \cref{tab:loss_comparison}. The results demonstrate that the proposed Spot loss and PSF loss significantly improve the performance. It is worth noting that Spot loss and PSF similarity loss can only be used together to maximize the performance of tolerance-aware optimization. The two losses play different roles respectively, the Spot loss is helpful to ensure that the overall imaging quality of the optics is not seriously degraded during the tolerance optimization, while the PSF loss is able to significantly improve the robustness of the optics to random tolerances on this basis. Without the constraint of Spot loss, PSF loss can lead to significant degradation in the imaging performance of the optical system, thereby impacting overall tolerance optimization. As shown in \cref{tab:loss_comparison}, the spot size is extremely large, indicating very low imaging quality.
\begin{figure}[t]
  \centering
   \includegraphics[width=1.0\linewidth]{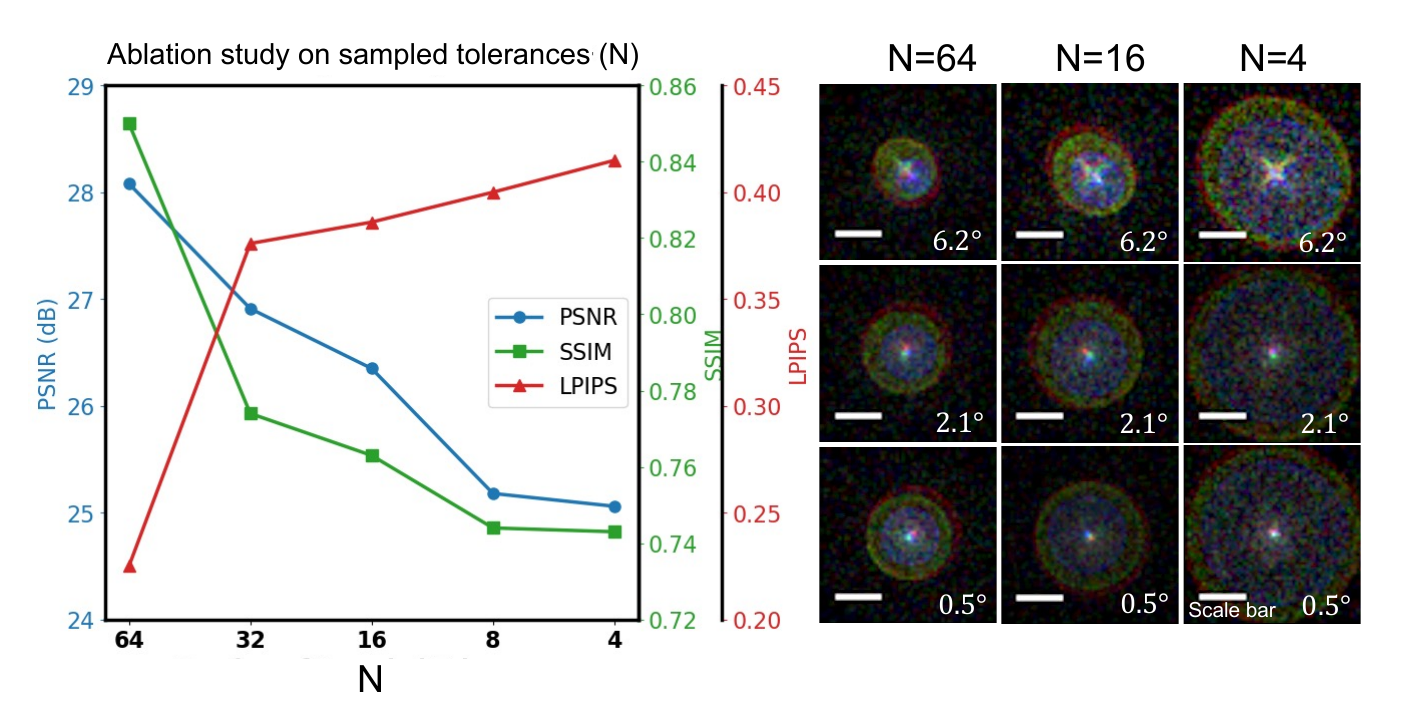}

   \caption{Ablation study on the number of sampled tolerances pattern in every iteration. Average PSNR, SSIM, and LPIPS of 100 times random tolerances test for different sampling numbers are shown, along with the PSFs comparison. Scale bar: $15\mu m$ .}
   \label{fig:perb_num}
\end{figure}

\begin{table}
\centering
\caption{Ablation study on Spot loss and PSF loss, each incorporating basic image quality loss. The results represent the averages from 100 random sampling tolerance experiments (excluding Spot Size).}
\begin{tabular}{c|c c c c}
\hline
 & - & $\mathcal{L}_{\text{Spot}}$ & $\mathcal{L}_{\text{PSF}}$ & $\mathcal{L}_{\text{Spot}}\&\mathcal{L}_{\text{PSF}}$ \\
\hline
PSNR$\uparrow$ & 24.83 & 26.63 & 20.49 & \bf{28.08} \\
SSIM$\uparrow$ & 0.738 & 0.782 & 0.601 & \bf{0.850} \\
LPIPS$\downarrow$ & 0.378 & 0.248 & 0.529 & \bf{0.225} \\
Spot Size ($\mu m$)$\downarrow$ & 52.6 & \bf{12.9} & 637.9 & 38.7 \\
\hline
\end{tabular}
\label{tab:loss_comparison}
\end{table}

\subsection{Number of Sampled Tolerance Patterns}
In the tolerance-aware optimization process, the number of sampled tolerance patterns is a critical hyperparameter that significantly impacts the performance. If the number is too low, the optimization may stuck in local optima. However, increasing the sampled number leads to a higher consumption of GPU memory. Therefore, selecting an appropriate sample number is of great importance to balance the trade-off between optimization effectiveness and memory usage. Therefore, we employ different number of sampled tolerance pattern and conduct quantitative analysis to determine a reasonable lower limit, see in \cref{fig:perb_num}. This experiment significantly streamlines subsequent research efforts.

\section{Conclusion}
In this paper, we present a novel end-to-end approach to bridge the gap between design and manufacturing in deep optics. By integrating tolerance modeling into the ray tracing process, we achieve effective tolerance-aware optimization, resulting in a more robust and complete deep optics design flow. Our analysis shows significant improvements in robustness, demonstrated through simulations and physical experiments. We believe this work effectively closes the divide between design and manufacturing, unlocking the potential of deep optics for mass production.

{
    \small
    \bibliographystyle{ieeenat_fullname}
    \bibliography{main}
}

\clearpage

\appendix
\renewcommand\thefigure{A\arabic{figure}}
\renewcommand\thetable{A\arabic{table}}  
\renewcommand\theequation{A\arabic{equation}}
\setcounter{section}{0}
\setcounter{equation}{0}
\setcounter{table}{0}
\setcounter{figure}{0}

\setcounter{page}{1}
\maketitlesupplementary

\section{Tolerances sampler}
\label{sec:supp}
In this section, we present more implementation details of our tolerance sampler and its integration into the ray tracing. We model four common kinds of tolerances: decentration, tilt, central thickness, and curvature errors, and the specific tolerance ranges we used see \cref{tab:tolr_range}. Considering the perturbations arising from tolerances in the design parameters, we can express the parameters as follows:
\begin{equation}
\theta_{real} = \theta_{ideal} + \theta_{\Delta},
\end{equation}
where $\theta_{ideal}$ is the design parameters from pretrained without considering tolerances, $\theta_{\Delta}$ is the perturbations come from random tolerances, $\theta_{real}$ is the optics parameters after fabrication. 

We firstly random sample tolerances from given ranges, thus, $\theta_{\Delta}$ obeys the normal distribution, $\theta_{\Delta} \sim \mathcal{N}(0, \frac{max^{2}}{9})$, so that the range of tolerances is approximately 3 times the standard deviation of the normal distribution, \eg, for decentration, $max=0.04mm$, and we clamp the tolerances range if sampled tolerance values exceed the $max$. Secondly, for each individual lens (or double glued structure) in the lens system, we independently sample four tolerances.

Given that we are conducting ray tracing on a surface-by-surface basis, we convert the sampled per-lens tolerances into the necessary spatial transformations and curvature offsets for each surface. First, we categorize the tolerances into two types: spatial transformations, which include decentration, tilt, and central thickness error, and curvature errors, which can be implemented as offsets in the ray tracing process. Notably, both decentration and central thickness errors are consolidated into a single translation of the lens surfaces, expressed as translation vector $\boldsymbol{T} = [\Delta X, \Delta Y, \Delta Z]^T$.  While tilt is represented as a rotation of the lens surfaces, expressed as Rotation matrix, $\boldsymbol{R} = \boldsymbol{R}_z(\gamma)\cdot \boldsymbol{R}_y(\beta)\cdot \boldsymbol{R}_x(\alpha)$, more specific:
\begin{align}
\boldsymbol{R}_x(\alpha) &= \begin{bmatrix}
1 & 0 & 0 \\
0 & \cos{\alpha} & \sin{\alpha} \\
0 & \sin{\alpha} & \cos{\alpha} \\
\end{bmatrix}, \\
\boldsymbol{R}_y(\beta) &= \begin{bmatrix}
\cos{\beta} & 0 & \sin{\beta} \\
0 & 1 & 0 \\
-\sin{\beta} & 0 & \cos{\beta}
\end{bmatrix}, \\
\boldsymbol{R}_z(\gamma) &= \begin{bmatrix}
\cos{\gamma} & -\sin{\gamma} & 0 \\
\sin{\gamma} & \cos{\gamma} & 0 \\
0 & 0 & 1
\end{bmatrix}.
\end{align}
where $\alpha, \beta, \gamma$ are sampled tilt degrees from normal distribution, around three axes.

\begin{table}
\centering
\caption{Specific tolerance ranges use in our paper. Decentration and central thickness error tolerances are in \emph{millimeters}, curvature error is in \emph{percentrage} and tilt is in \emph{degree}.}
\begin{tabular}{c|c}
\hline
& TOLR Range \\
\hline
Decenteration & $(-0.04, 0.04)$ \\
Tilt & $(-0.05, 0.05)$ \\
Central thickness error & $(-0.04, 0.04)$ \\
Curvature error & $(-0.3\%, +0.3\%)$ \\
\hline
\end{tabular}
\label{tab:tolr_range}
\end{table}

As we convert the surface transformations into equivalent coordinate system transformations, we need to invert the signs of the formulas above. In the ray tracing process, we first apply the coordinate system transformations in the order of translation followed by rotation. We then calculate the intersection point between the ray and the surface, incorporating the curvature offset to finalize the application of Snell's law. Finally, we revert the coordinate system back to its original configuration.

\section{Deep optics pipeline}
In this section, we present more implementation details for deep optics pipeline in both training and testing.

\subsection{Rendering by PSF map}
\begin{figure*}[tb] \centering
  \includegraphics[width=0.9\textwidth,height=0.32\textwidth]{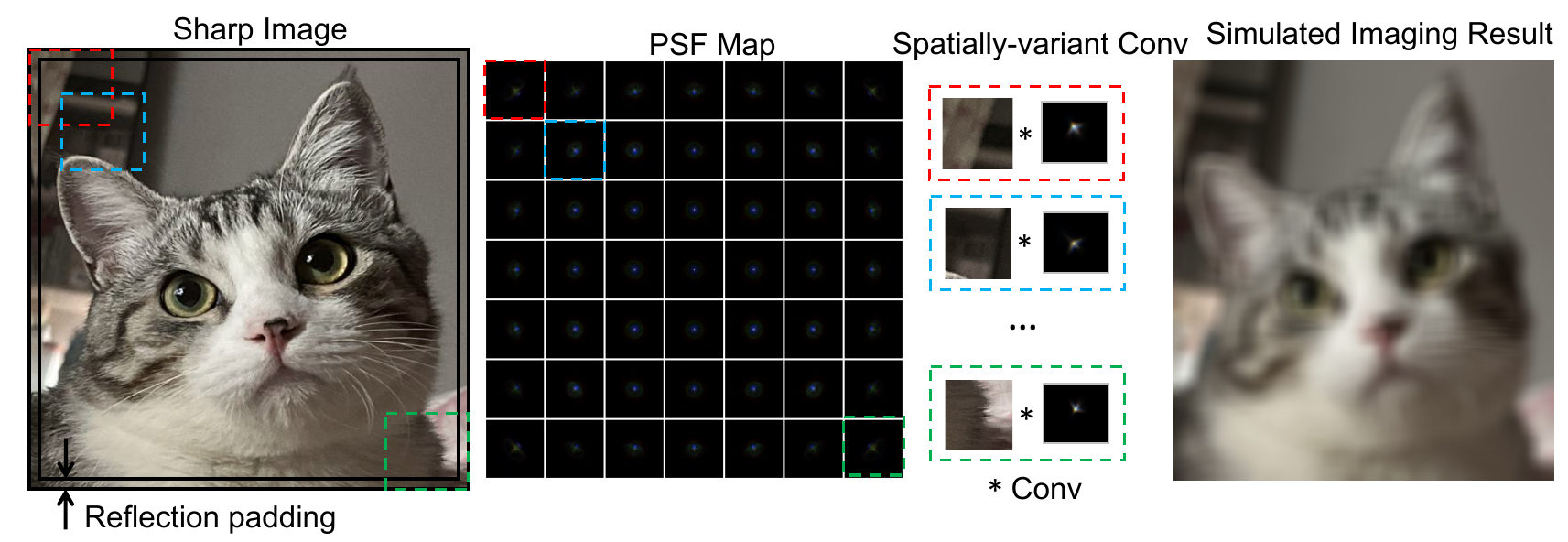}
  \caption{Initially, the original image is populated with reflections. Subsequently, images at various field-of-view positions within the complete large image are convolved using the point spread function (PSF) corresponding to each field-of-view as the convolution kernel. Finally, the convolved images are integrated to produce a cohesive composite.} 
  \label{fig:spatially-conv}
\end{figure*}

Rather than rendering the imaging result pixel by pixel, we first generate the Point Spread Function (PSF) Map through ray tracing. Once the PSF Map is obtained, we compute the camera imaging result using spatially-variant convolution between the PSF Map and the sharp image, illustrated by \cref{fig:spatially-conv}.

\begin{figure*}[tb] \centering
  \includegraphics[width=0.8\textwidth,height=0.32\textwidth]{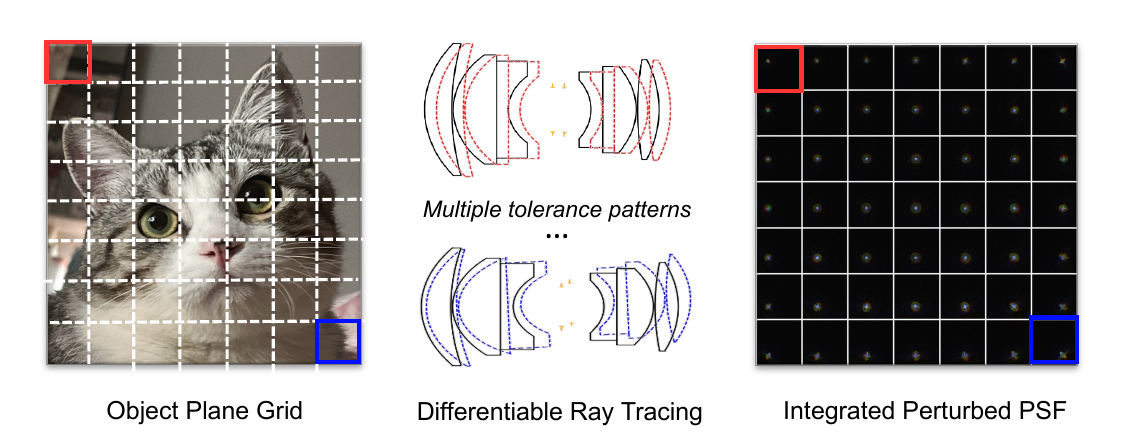}
  \caption{For each local field of view (FoV), a random tolerance pattern is sampled, and the local Point Spread Function (PSF) is obtained through differentiable ray tracing. These local PSFs are then stitched together to create a PSF map that incorporates multiple tolerance patterns.} 
  \label{fig:combine_PSF}
\end{figure*}

In this paper, we adopt the differentiable PSF map method from \cite{yang2024curriculum}. In our experiments, we utilize $51 \times 51$ PSF kernel size and $8\times 8$ PSF grid map.

\subsection{Tolerance optimization and evaluation}
\textbf{Tolerance optimization.} During the tolerance optimization process, we need to simultaneously sample various tolerance modes and perform ray tracing. To mitigate computational overhead, we conduct local ray tracing within a distinct field of view for each sampled tolerance mode, allowing us to obtain the PSF at specific local field of view angles. In our experiments, we sample 64 tolerance patterns, rendering a local PSF for each pattern. Finally, we combine all 64 local PSFs into an integrated PSF map that encapsulates the effects of all tolerance patterns, demonstrated in \cref{fig:combine_PSF}. It is important to note that we utilize resized images ($256 \times 256$) from the DIV2K dataset \cite{agustsson2017ntire} during the training process to mitigate memory overhead, batch size is set in 64. During tolerance optimization, we employ each localized PSF from the PSF map alongside a $256 \times 256$ image for convolution to obtain simulated imaging results. Finally, we add $1\%$ Gaussian noise to simulate sensor noise, as shown in \cref{fig:pipeline}.

\vspace{0.8em}
\noindent\textbf{Evaluation.} During evaluation, in contrast to the training phase, we directly utilize $2048 \times 2048$ images along with a PSF map obtained from ray tracing, applying spatially-variant convolution to generate simulated imaging results, batch size is set in 2. We sampled randomized tolerances multiple times, and for each sampled tolerance pattern, we computed the PSNR, SSIM and LPIPS, across the entire test dataset. 

\begin{figure*}[tb] \centering
  \includegraphics[width=\textwidth,height=0.64\textwidth]{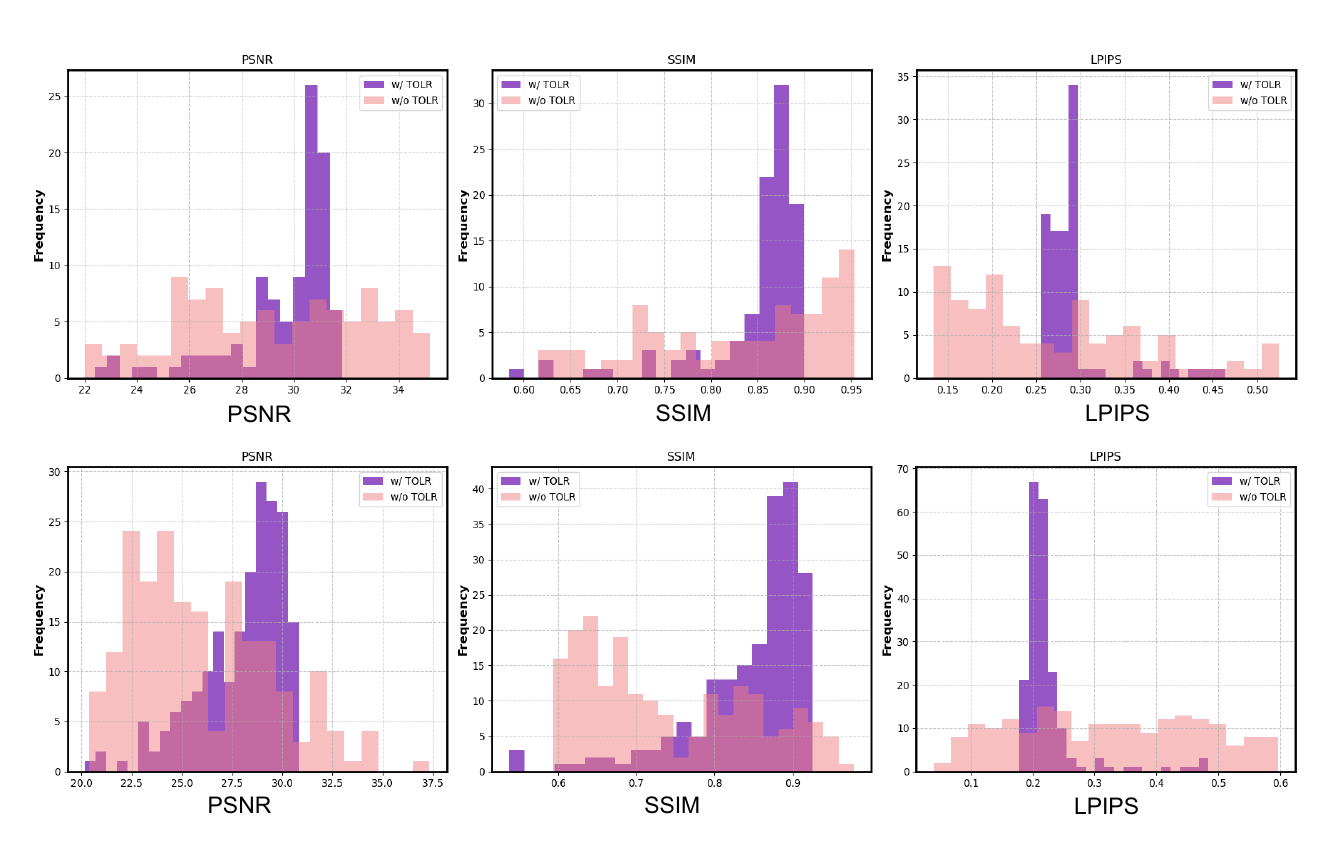}
  \caption{Top: 100 times of random tolerances test for Lens1. Bottom: 200 times of random tolerances test for Lens2. The frequency distribution of the overall test indicates that the deep optics design, following tolerance-aware optimization, is significantly less impacted by tolerances.} 
  \label{fig:distribution}
\end{figure*}

We visualize the frequency histograms of the complete test results for Lens 1 and Lens 2, see \cref{fig:distribution}.
We provide a visual flipping comparison in the local web page, please refer to the file \textit{visualization.html} in the supplement.

\section{Lens1 and Lens2 parameters}
\begin{figure*}[tb]
\centering
  \includegraphics[width=0.7\textwidth,height=0.25\textwidth]{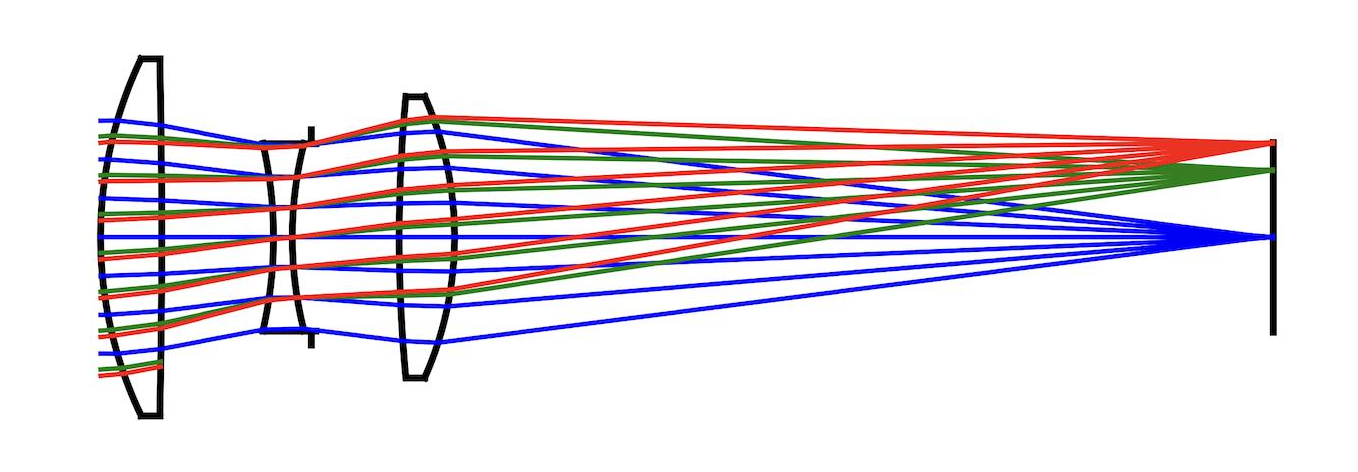}
  \caption{Lens1 layout.} 
  \label{fig:lens1_layout}
\end{figure*}

\begin{figure*}[tb]
\centering
  \includegraphics[width=0.7\textwidth,height=0.25\textwidth]{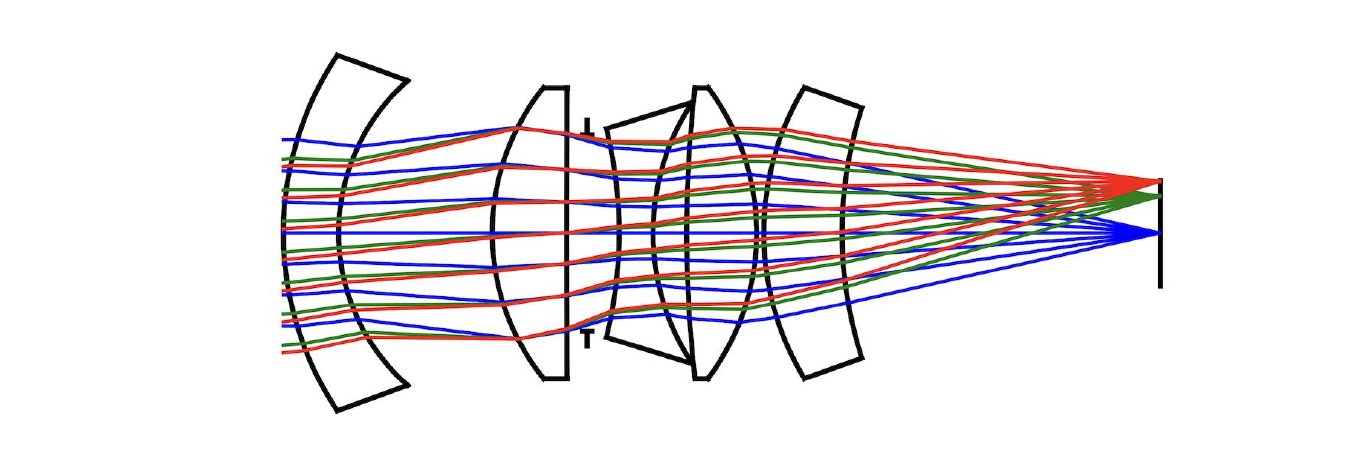}
  \caption{Lens2 layout.} 
  \label{fig:lens2_layout}
\end{figure*}

In this section, we list all parameters of lens1 and lens2 in \cref{tab:lens1_param} and \cref{tab:lens2_param}, and the layouts for the two lenses are as shown in \cref{fig:lens1_layout} and \cref{fig:lens2_layout}.

\begin{table*}
\centering
\caption{The lens1 parameters used in our paper.}
\label{tab:lens1_param}
\begin{tabular}{cccccc}
\toprule
Surface No. & Radius (mm) & Distance (mm) & Diameter (mm) & $n_d$ & $\nu_d$ \\
\toprule
OBJ & INFINITY & 1000.0 & INFINITY & AIR &  \\
1 & 22.01 & 3.26 & 19.0 & 1.620410 & 60.323649 \\
2 & -435.76 & 6.01 & 19.0 & AIR & \\
3 & -22.21 & 1.0 & 10.0 & 1.620040 & 36.376491 \\
4 & 20.29 & 1.0 & 10.0 & AIR & \\
STO & INFINITY & 4.75 & 10.0 & AIR & \\
5 & 79.68 & 2.95 & 15.0 & 1.620410 & 60.323649 \\
6 & -18.40 & 41.55 & 15.0 & AIR & \\
IMA & INFINITY & - & 10.14 & AIR & \\
\bottomrule
\end{tabular}
\end{table*}

\begin{table*}[htp]
\centering
\caption{The lens2 parameters used in our paper.}
\label{tab:lens2_param}
\begin{tabular}{cccccc}
\toprule
Surface No. & Radius (mm) & Distance (mm) & Diameter (mm) & $n_d$ & $\nu_d$ \\
\toprule
OBJ & INFINITY & 2000.0 & INFINITY & AIR &  \\
1 & 37.789 & 6.544 & 42.0 & 1.6400 & 60.20 \\
2 & 23.941 & 18.170 & 35.984 & AIR & \\
3 & 27.340 & 8.800 & 34.400 & 1.7880 & 47.49 \\
4 & 2541.820 & 2.432 & 34.400 & AIR & \\
STO & INFINITY & 3.800 & 23.200 & AIR & \\
5 & -50.594 & 4.000 & 24.640 & 1.7552 & 27.53 \\
6 & 28.004 & 3.994 & 30.862 & AIR & \\
7 & 149.795 & 8.285 & 34.400 &  1.6400 & 60.20 \\
8 & -28.474 & 0.832 & 34.400 & AIR & \\
9 & 33.206 & 9.275 & 34.400 &  1.6584 & 50.85 \\
10 & 47.804 & 35.074 & 29.560 & AIR & \\
IMA & INFINITY & - & 12.454 & AIR & \\
\bottomrule
\end{tabular}
\end{table*}

\end{document}